\documentclass{article}

\usepackage{arxiv}
\usepackage{textcomp}

\usepackage[utf8]{inputenc} % allow utf-8 input
\usepackage[T1]{fontenc}    % use 8-bit T1 fonts
\usepackage{hyperref}       % hyperlinks
\usepackage{url}            % simple URL typesetting
\usepackage{booktabs}       % professional-quality tables
\usepackage{amsfonts}       % blackboard math symbols
\usepackage{nicefrac}       % compact symbols for 1/2, etc.
\usepackage{microtype}      % microtypography
\usepackage{lipsum}
\usepackage{graphicx}
\usepackage{multirow}
\usepackage{floatrow}
\newfloatcommand{capbtabbox}{table}[][\FBwidth]
\graphicspath{ {./images/} }

\usepackage{multirow}

\usepackage{makecell}
\usepackage{xcolor}
\usepackage{colortbl}
\usepackage{url}
\usepackage{listings}
\usepackage{enumitem}

\usepackage[font=scriptsize]{subfig}

\definecolor{codegreen}{rgb}{0,0.5,0}
\definecolor{codeblue}{rgb}{0.25,0.5,0.5}
\definecolor{codegray}{rgb}{0.6,0.6,0.6}
\usepackage{color}

\definecolor{dkgreen}{rgb}{0,0.6,0}
\definecolor{gray}{rgb}{0.5,0.5,0.5}
\definecolor{mauve}{rgb}{0.58,0,0.82}

\usepackage{bm}

\usepackage{authblk}

\begin{document}

\title{1st Place Solution for ICDAR 2021 Competition on Mathematical Formula Detection}

\author[1]{Yuxiang Zhong}
\author[1]{Xianbiao Qi}
\author[1]{Shanjun Li}
\author[1]{Dengyi Gu}
\author[1]{Yihao Chen}
\author[1]{Peiyang Ning}
\author[1]{Rong Xiao}

%\author{
% Yuxiang Zhong, \And Xianbiao Qi, \And Shanjun Li, \And Dengyi Gu, \And Yihao Chen, \And Peiyang Ning, \And Rong Xiao
%}
\affil{ Visual Computing Group (VCGroup), Ping An Property \& Casualty Insurance Company.}
\thanks{Xianbiao Qi is the corresponding author. If you have any questions or concerns about the implementation details, please do not hesitate to contact 1701210384@pku.edu.cn or qixianbiao@gmail.com.}

\maketitle

\begin{abstract}
In this technical report, we present our 1st place solution for the ICDAR 2021 competition on mathematical formula detection (MFD)\footnote{\url{http://transcriptorium.eu/~htrcontest/MathsICDAR2021/}}. The MFD task has three key challenges, i.e.\  a large scale span, large variation of the ratio between height and width, and rich character set and mathematical expressions. 
Considering these challenges, we used Generalized Focal Loss (GFL), an anchor-free method, instead of the anchor-based method, and prove the Adaptive Training Sampling Strategy (ATSS) and proper Feature Pyramid Network (FPN) can well solve the important issue of scale variation. Meanwhile, we also found some tricks, e.g., Deformable Convolution Network (DCN), SyncBN, and Weighted Box Fusion (WBF), were effective in MFD task. \bf{Our method was ranked 1st place in the final 15 teams}. 
\end{abstract}

%In our method, we only use a detection pipeline to accomplish this task. We adopt ResNeSt~\cite{ResNestzhang2020resnest}, which is a SOTA feature extractor, as our backbone. And our detection algorithm is customized baesd on GFL~\cite{GFLli2020generalized}, a robust single stage detector.
%Finally, Weighted-Box-Fusion~\cite{WBFsolovyev2019weighted} is employed to model ensemble. 

% keywords can be removed

\section{Introduction}
Digitization of document images is vital for enabling searching in massive collections of digitized printed scientific documents. Different from common text content, general OCR software fails to process mathematical formulas. Usually, a specific mathematical expression recognition is needed for mathematical formulas. To do this, we have to detect the regions of mathematical formulas. The ICDAR 2021 competition on mathematical formula detection (MFD) is targeting this problem. In the MFD task, there are two forms of formulas, i.e.\ embedded and isolated.

Different from general object detection~\cite{girshick2015fast,redmon2016you,liu2016ssd,FASTERRCNNren2015faster,cai2018cascade}, the MFD task~\cite{mahdavi2019icdar,chu2013mathematical,gao2017deep} has three key characteristics, i.e.\ large scale variance especially between isolated and embedded formulas, large variation of the ratio between width and height, and rich character set and mathematical expressions. In this paper, we fully consider the features of the MFD task, and develop an empirical solution that is ranked 1st place in the ICDAR 2021 competition on mathematical formula detection. In our solution, we used an anchor-free framework, Generalized Focal Loss (GFL)~\cite{GFLli2020generalized}, instead of the anchor-based model~\cite{FASTERRCNNren2015faster} because the proposal generated 
by the latter is hard to cover different scales and different ratios between width and height. Meanwhile, we employed an adaptive training sample selection (ATSS)~\cite{ATSSzhang2020bridging} to balance the sampled positive points on each formula instance. In addition, we leveraged the feature pyramid network (FPN)~\cite{FPNlin2017feature}, the large input resolution, and the deformable convolutional network (DCN)~\cite{dai2017deformable} because these techniques could benefit large formula detection or small formula detection, or both. Finally, some other tricks, such as ResNeSt~\cite{ResNestzhang2020resnest}, SyncBN~\cite{zhang2018context}, the large batch size and the weighted box fusion (WBF)~\cite{solovyev2019weighted}, Ranger~\cite{lessw2019ranger} optimizer, were adopted in our solution.

%1. 我们为什么采用one-stage GFL
%2. ATSS是重要的， 
%3. FPN，large resolution, DCN 是重要的
%Furthermore, we also use NeSt, SyBN, LBS WBF等

%Searching in massive collections of digitized printed scientific documents with queries that are mathematical expressions is a research area scarcely explored. To address this problem, a crucial first step involves the detection of regions that may contain mathematical expressions. The ICDAR 2021 competition on mathematical formula detection aims to tackle this problem. 

%The IBEM data set is provided as the official data In this competition, and Intersection-over-Union (IoU) metric is used for evaluation. The proposed IBEM dataset consists of a subset of documents from the KDD Cup 2003 competition dataset. The IBEM dataset has been automatically generated by processing the LaTeX version of STEM papers available in the KDD Cup dataset. The ground truth extracted contains information about the position of mathematical expressions at page level and their LaTeX definition. Given that the LaTeX definitions of these mathematical formulas play no role in the detection process, the competition provide a simplified version of the IBEM dataset with only the relevant information. The mathematical formulas that appear in STEM documents, and therefore in the IBEM dataset, are either embbeded along the lines of the text or isolated. For each mathematical expression, either embedded or isolated, information on the exact position of its minimum bounding box will be provided.

\begin{figure}[h] % picture
    \centering
    \includegraphics[width=1.0\textwidth]{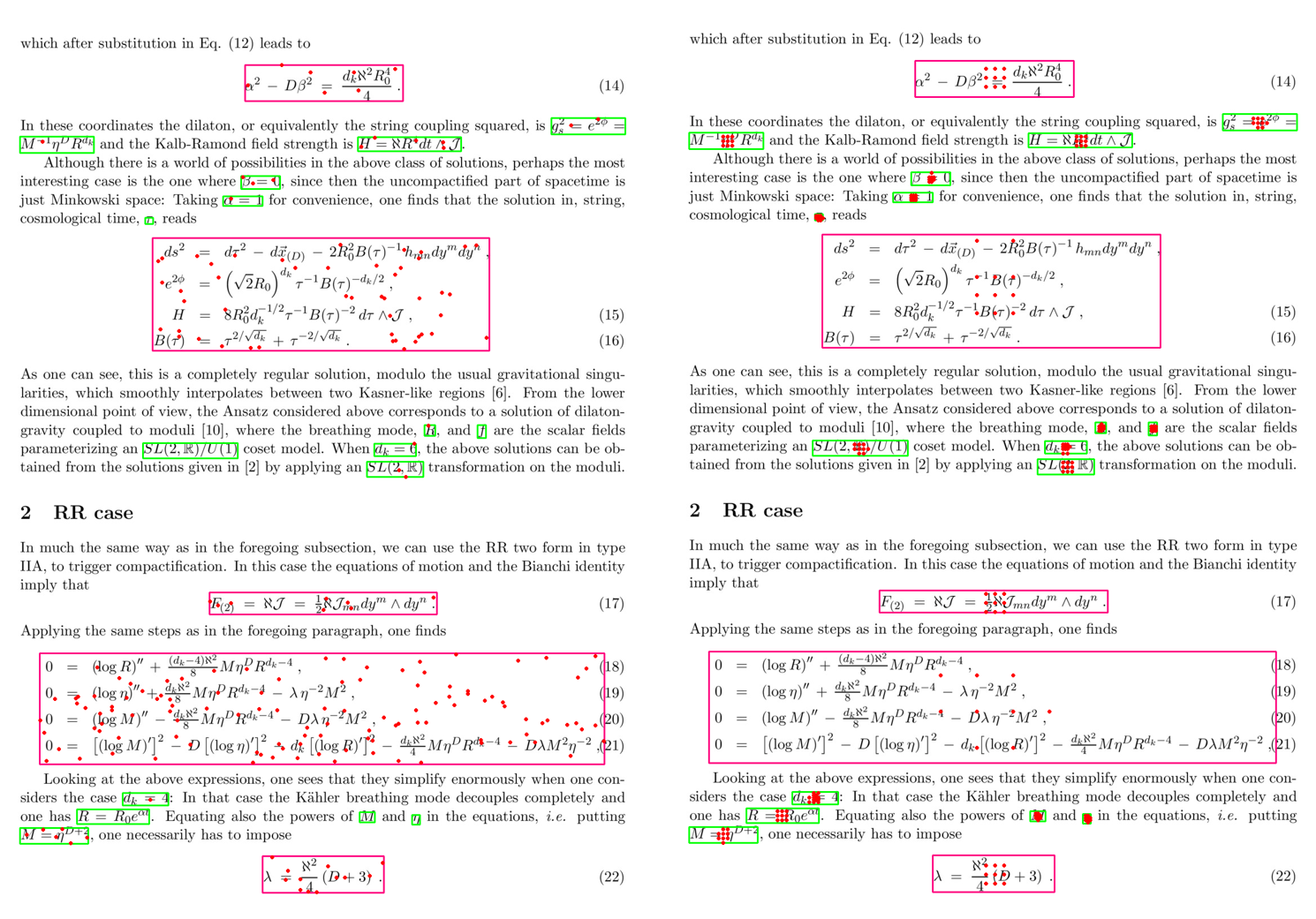}
    \caption{Comparison of random sampling strategy and ATSS sampling. Left. positive samples with random sampling strategy; Right. positive samples with ATSS.}
    \label{fig:sampling_strategy}
\end{figure}

%\begin{figure}[h] % picture
%    \centering
%    \includegraphics[width=1.0\textwidth]{image/challenge_analysis.png}
%    \caption{(left) Dense tiny formula in Document; (right) Huge scale range between embedded and isolate formula}
%    \label{fig:challenge_analysis}
%\end{figure}

\section{Challenge}
Compared to general object detection in natural images or scene text detection in documents, mathematical formula detection in this task will encounter some specific difficulties. 
As shown in Figure~\ref{fig:sampling_strategy}, these challenges include
\begin{itemize}[leftmargin=.2in]
    \item a large scale span between embedded formula and isolated formula. The areas of some large isolated formulas are hundreds of times larger than that of some embedded formulas;
    \item huge variation of the ratio between height and width. As shown in Figure~\ref{fig:sampling_strategy}, the ratios range from around 0.5 to 12;
    \item rich character set and mathematical expressions. There are hundreds of mathematical characters, and their combinations are extremely rich.
\end{itemize}
The large scale span of formulas and the huge variation of the ratio between height and width will lead to a classical scale problem in object detection. To be specific, as shown in the left of Figure ~\ref{fig:sampling_strategy}, isolated formulas with large area will obtain large number of positive samples, but small embedded formulas only have a few or even none of positive samples under the random sampling strategy.
In addition, large scale span will bring in a challenge to the head of the detector. To deal with this issue, attention needs to be paid to the choice of the pyramid of the detector head. In the following section, we will describe methods targeting these two challenges.

%Firstly, there is a large scale span between embedded and isolated formula, as shown in Figure~\ref{fig:challenge_analysis}. It may lead to the classical scale problem in object detection and sampling imbalance with random positive sampling strategy~\cite{FASTERRCNNren2015faster}. Moreover, a great number of embedded formula will appear to the same page normally, can be seen in Figure~\ref{fig:challenge_analysis}(left), and it introduces another problem with tiny dense object detection.

%The rest of the paper is organized as follows. Firstly, we will introduce the methods of this competition in section 2. Some experiment results will be presented in section 3. And finally, section 4 concludes the paper. 

%On classification branch, the quality of bounding box is introduced in conjunction with the classification score. On regression branch, the inflexible Dirac  delta distribution is replaced by general distribution, to get more precise location.

\section{Method}
\label{sec:headings}
We treated this task as a pure object detection task, and adopted Generalized Focal Loss (GFL~\cite{GFLli2020generalized}) framework as our baseline. 
GFL is built on a Fully Convolutional One-Stage (FCOS)~\cite{FCOStian2019fcos} detector that is a robust anchor-free method. Based on the up-mentioned challenges, we reckon the anchor-free method is better than the anchor-based methods (e.g., Faster R-CNN~\cite{FASTERRCNNren2015faster}, Cascade R-CNN~\cite{cai2018cascade}) in this task. 
In the following, we will describe two key modules in detail that greatly improved our solution.

%\textbf{ResNeSt.} ResNeSt is adopted as our backbone network.

%\textbf{FCOS.} Fully Convolutional One-Stage Detector, a robust and scalability anchor-free method. It is employed to our base detector.

\textbf{ATSS.} Adaptive Training Sample Selection~\cite{ATSSzhang2020bridging} is an effective sampling strategy to alleviate the complex artificial design of sampling ratio between positive and negative, and overcomes the sampling imbalance disadvantage of random strategy. As shown in the left of the Figure~\ref{fig:sampling_strategy}, the results of random sampling strategy obey an uniform distribution according to the area of positive samples. It means that larger formula will obtain more positive samples, while some tiny formulas will only get a few or even zero positive samples. The sampling frequency of these tiny embedded formulas is much lower than the large isolated formulas. Obviously, there is a sampling imbalance between tiny and large formulas with random sampling strategy. On the contrary, when using ATSS, the number of positive samples has no relationship with the area of instance. Each instance will get the same number of positive samples. As shown in the right of the Figure~\ref{fig:sampling_strategy}, ATSS can effectively eliminate the impact of imbalanced sampling.

\begin{figure}[h] % picture
    \centering
    \includegraphics[width=0.80\textwidth]{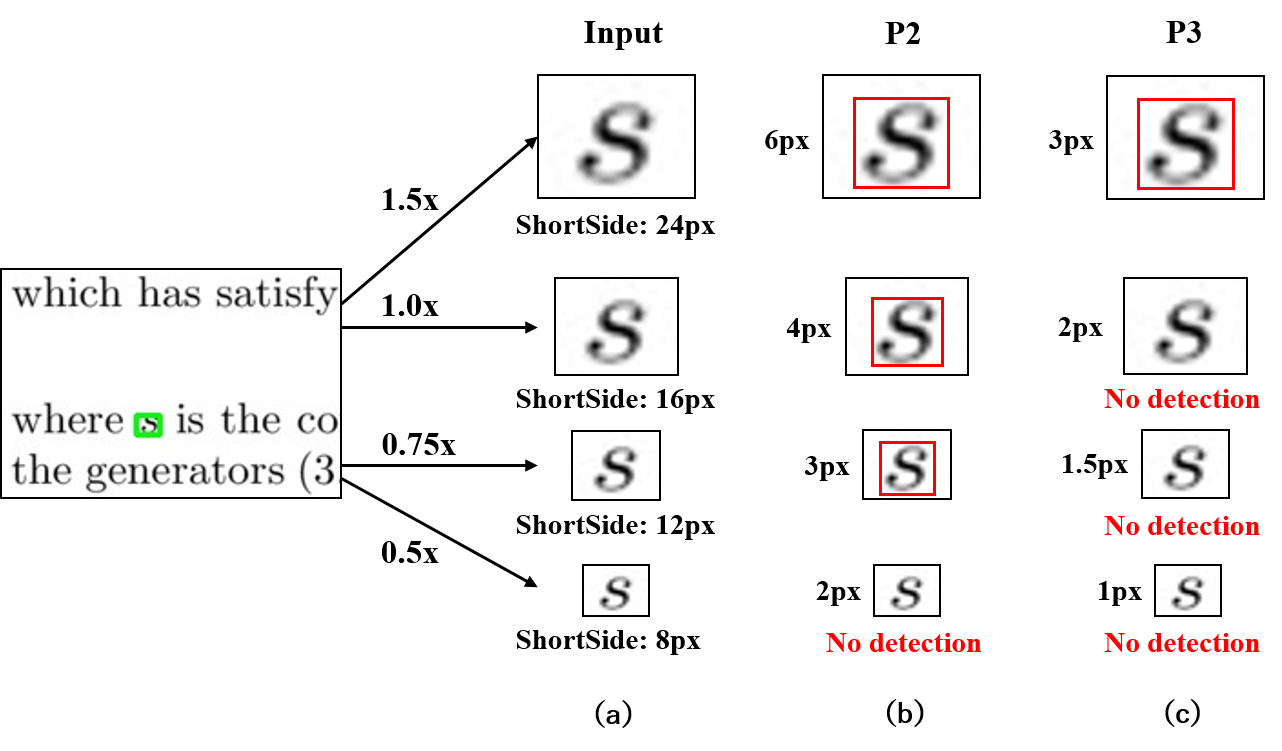}
    \caption{Illustration of the importantce of the FPN on the MFD task. (a) Input size; (b) Corresponding size in P2; (c) Corresponding size in P3. Under some small scales, some positive samples will be missed.}
    \label{fig:fpn_analysis}
\end{figure}

\textbf{FPN.} Feature Pyramid Network (FPN)~\cite{FPNlin2017feature} is to explicitly address the problem of large scale span. The MFD task contains a large number of extremely small formulas, which bring great challenges to our model. As illustrated in Figure~\ref{fig:fpn_analysis}, for such a single extremely small character formula, their short sides are usually around 16 pixels. We observe that for any layer of FPN, the limit of the detector is 3 pixels. This means that if we use the default FPN (3-7), the short side of the formula needs to be at least 24 pixels to be detected. Obviously, there are many small embedded formulas that do not satisfy this condition. So we change the selection of FPN level to (2-6) so that our model can overcome this defect. 
%Judging from the results, our score in embeded formula has increased significantly.
%\textbf{GFL.} We utilize Generalized Focal Loss as loss function of our detector. 

\section{Experiment}
\label{sec:others}
In this section, we will describe the implementation of our mathematical formula detection system in detail.

{\bf{Dataset.}} Our used data is the official IBEM dataset. The IBEM dataset consists of 600 documents with a total number of 8,273 pages. There are 58,834 isolated and 260,323 embedded expressions. Each document contains approximately an average of 14 pages, 98 isolated mathematical expressions, and 434 embedded mathematical expressions. Each document is separated into individual pages with the corresponding ground truth. It is unknown how the organizers make a change to the original IBEM data set. 
Finally, there are 7,244 images with labels for training, and 4,028 images without a label for final evaluation. Only the provide training data is used for training.

%For once end-to-end inference, one table image will forward PSENet, table structure MASTER and text line recognition MASTER. Box assignment(matching rules) and some HTML format rules will be used to get the final HTML code.

{\bf{Implementation Details.}} Our baseline model is GFL with ResNet101~\cite{RESNEThe2016deep} as the backbone. In the training of the baseline model, 8 Tesla V100 GPUs are used with batch size 2 in each GPU. The input image is resized equally to $1447\times 2048$ with a keeping ratio. An $800\times 800$ region is cropped from the original image. We use FPN (3-7) in default. In the baseline model, we use Adam~\cite{kingma2014adam} optimizer.

%RandomFilp and RandomCrop are used for data augmentation. 
%A $1440\times 1600$ region is cropped from each image. 

In the inference stage, we resize the test image to $1583\times 2048$ due to the resolution distribution of the test dataset. Flip augment is turned on. For the post-processing, Non-Maximum Suppression (NMS)~\cite{rothe2014non} with 0.6 IoU threshold is applied to filter the redundant box. All models are trained based on MMDetection~\cite{MMDetectionchen2019mmdetection} toolbox.

%In the final model ensemble schedules, we select two best performance model which are initial with different random seed. IoU threshold of NMS is set 0.4 in Weighted-Box-Fusion.

\subsection{Ablation Study}
We have conducted many attempts in this competition. Some effective tricks are listed below. We list these tricks according to the timeline we tried them.
%In this subsection, we take ResNet50~\cite{RESNEThe2016deep}+GFL default config in MMDetection and random crop size $800\times 800$ as our baseline. We will discuss some useful tricks, but ignore some unsuccessful attempts. The result of ablation studies is shown in Table~\ref{ablation_study} 

{\bf{Double Training Epoch. (DTE)}} Initially, We train 12 epochs. Then, we extend the training epoch to 24, and get a substantial score gain. Therefore, we further train 36 epochs, the result has no obvious change. It means that model is under-fitting with 12 epochs, and suitable with 24 epochs. 

{\bf{Large Crop Size.}} Considering that a small crop size will cause the long-isolated formula to be truncated sometimes. We decide to expand the crop size to $1440\times 1600$.

{\bf{Random Flip}} is used at the training phase with a 0.5 ratios. In the inference phase, the image is also flipped as the input. The result is attained by mapping back the flipped boxes. The final result will be obtained by NMS. 

{\bf{ResNeSt}}~\cite{ResNestzhang2020resnest} is used as our backbone to replace ResNet. Here, we use ResNeSt101 as the backbone.

{\bf{Synchronized Batch Normalization (SyncBN)}}~\cite{zhang2018context} is an effective batch normalization approach that is more suitable for the situation that the batch size is relatively small on each GPU graphics card. 

{\bf{Deformable Convolution Network (DCN)}}~\cite{dai2017deformable}. There are some discrete and large isolated formulas in the data set. To detect such formulas, a large receipt filed is needed. Deformable convolution is an effective way to extend the receptive field. We replace all convolutions with deformable convolution layers from c3 to c5 in ResNeSt101. %As expected, the our model's performance on isolated formula has been greatly improved. 

{\bf{Larger Batch Size (LBS).}} We expand batch size from 2 to 3 each GPU by utilizing the memory consume, such as mixed precision and replaced ReLU.

{\bf{Ranger}}~\cite{lessw2019ranger} is a synergistic optimizer combining RAdam (Rectified Adam)~\cite{liu2019radam}, LookAhead~\cite{zhang2019lookahead}, and GC (gradient centralization)~\cite{yong2020gradient}. 
%We observe  that Ranger optimizer shows a better performance than Adam and SGD in this competition. 

{\bf{RegMax}} is a hyperparameter of GFL. It represents the largest position regression offset in each layer of FPN. In order to deal with the slim isolated formula, we increase the default 16 to 24. 

%As result shows, the score of isolated formula get a great growth.

{\bf{FPN Selection.}} We change the FPN (3-7) to FPN (2-6).

{\bf{Weight Box Fusion} (WBF)}~\cite{solovyev2019weighted} is used for model ensemble. It adopts an Expectation-Maximization algorithm to fuse the final bounding boxes. We select two models with the best performance which are trained initially with different random seeds. The IoU threshold of NMS is set to be 0.4 in WBF.

\begin{table}[]
\footnotesize
\begin{tabular}{cccccccccccccc}
 \multicolumn{1}{@{\ }l@{\ }}{DTE} & \multicolumn{1}{@{\ }l@{\ }}{LCS} & \multicolumn{1}{@{\ }l@{\ }}{Flip} & \multicolumn{1}{@{\ }l@{\ }}{NeSt} & \multicolumn{1}{@{\ }l@{\ }}{SyBN} & \multicolumn{1}{@{\ }l@{\ }}{DCN} & \multicolumn{1}{@{\ }l@{\ }}{LBS} & \multicolumn{1}{@{\ }l@{\ }}{Ranger} & \multicolumn{1}{@{\ }l@{\ }}{
\begin{tabular}[c]{@{}c@{}}Reg\\ 24\end{tabular}
} & \multicolumn{1}{@{\ }l@{\ }}{
\begin{tabular}[c]{@{}c@{}}FPN\\ (2-6)\end{tabular}
} & 
WBF
&
\multicolumn{1}{c}{
\begin{tabular}[c]{@{}c@{}}F1-score\\ Embedded\end{tabular}
}                             & \multicolumn{1}{c}{
\begin{tabular}[c]{@{}c@{}}F1-score\\ Isolated\end{tabular}
}                            & \multicolumn{1}{c}{
\begin{tabular}[c]{@{}c@{}}F1-score\\ Total\end{tabular}
}                               \\ [0.3cm]
\hline

                                                   &                         &                          &     &                        &                            &                         &                         &                            &                              &                              & \begin{tabular}[c]{@{}c@{}}89.88\\ p:91.34   r:88.47\end{tabular} & \begin{tabular}[c]{@{}c@{}}85.92\\ p:89.77   r:82.39\end{tabular} & \begin{tabular}[c]{@{}c@{}}\textbf{89.17}\\ p:91.02   r:87.39\end{tabular} \\ [0.3cm]
\hline
                              $\surd$                      &                         &                          &                             &            &                &                         &                         &                            &                              &                              & \begin{tabular}[c]{@{}c@{}}91.36\\ p:92.80   r:89.96\end{tabular} & \begin{tabular}[c]{@{}c@{}}86.29\\ p:90.93   r:82.10\end{tabular} & \begin{tabular}[c]{@{}c@{}}\textbf{90.45}\\ p:92.26   r:88.71\end{tabular} \\[0.3cm]
                             $\surd$                      & \textbf{$\surd$}             &                          &                             &                            &                &         &                         &                            &                              &                              & \begin{tabular}[c]{@{}c@{}}92.95\\ p:94.43   r:91.52\end{tabular} & \begin{tabular}[c]{@{}c@{}}87.84\\ p:92.04   r:84.01\end{tabular} & \begin{tabular}[c]{@{}c@{}}\textbf{92.03}\\ p:93.58   r:90.53\end{tabular} \\[0.3cm]
                             $\surd$                      & $\surd$                      & $\surd$                       &                             &                            &                         &           &              &                            &                              &                              & \begin{tabular}[c]{@{}c@{}}93.58\\ p:94.79   r:92.41\end{tabular} & \begin{tabular}[c]{@{}c@{}}88.38\\ p:92.29   r:84.78\end{tabular} & \begin{tabular}[c]{@{}c@{}}\textbf{92.66}\\ p:94.35   r:91.03\end{tabular} \\[0.3cm]
                             $\surd$                      & $\surd$                      & $\surd$                       & $\surd$                          &                            &    &                     &                         &                            &                              &                              & \begin{tabular}[c]{@{}c@{}}93.49\\ p:93.82   r:93.16\end{tabular} & \begin{tabular}[c]{@{}c@{}}91.33\\ p:93.56   r:89.20\end{tabular} & \begin{tabular}[c]{@{}c@{}}\textbf{93.12}\\ p:93.71   r:92.54\end{tabular} \\[0.3cm]
                             $\surd$                      & $\surd$                      & $\surd$                       & $\surd$                          & $\surd$                         &                         &                         &      &                      &                              &                              & \begin{tabular}[c]{@{}c@{}}93.81\\ p:94.00   r:93.63\end{tabular} & \begin{tabular}[c]{@{}c@{}}91.60\\ p:93.82   r:89.48\end{tabular} & \begin{tabular}[c]{@{}c@{}}\textbf{93.42}\\ p:93.97   r:92.89\end{tabular} \\[0.3cm]
                             $\surd$                      & $\surd$                      & $\surd$                       & $\surd$                          & $\surd$                         & $\surd$                      &                         &                            &         &                     &                              & \begin{tabular}[c]{@{}c@{}}94.33\\ p:95.62   r:93.07\end{tabular} & \begin{tabular}[c]{@{}c@{}}95.25\\ p:95.62   r:94.88\end{tabular} & \begin{tabular}[c]{@{}c@{}}\textbf{94.49}\\ p:95.62   r:93.39\end{tabular} \\[0.3cm]
                             $\surd$                      & $\surd$                      & $\surd$                       & $\surd$                          & $\surd$                         & $\surd$                      & $\surd$                      &                            &     &                         &                              & \begin{tabular}[c]{@{}c@{}}94.58\\ p:95.16   r:94.00\end{tabular} & \begin{tabular}[c]{@{}c@{}}95.60\\ p:95.97   r:95.23\end{tabular} & \begin{tabular}[c]{@{}c@{}}\textbf{94.76}\\ p:95.31   r:94.22\end{tabular} \\[0.3cm]
                             $\surd$                      & $\surd$                      & $\surd$                       & $\surd$                          & $\surd$                         & $\surd$                      & $\surd$                      & $\surd$                         &   &                           &                              & \begin{tabular}[c]{@{}c@{}}95.22\\ p:95.70   r:94.74\end{tabular} & \begin{tabular}[c]{@{}c@{}}95.81\\ p:96.12   r:95.51\end{tabular} & \begin{tabular}[c]{@{}c@{}}\textbf{95.33}\\ p:95.79   r:94.87\end{tabular} \\[0.3cm]
        $\surd$                      & $\surd$                      & $\surd$                       & $\surd$                          & $\surd$                         & $\surd$                      & $\surd$                      & $\surd$                         & $\surd$                           &         &                     & \begin{tabular}[c]{@{}c@{}}95.01\\ p:95.86   r:94.16\end{tabular} & \begin{tabular}[c]{@{}c@{}}97.28\\ p:97.18   r:97.38\end{tabular} & \begin{tabular}[c]{@{}c@{}}\textbf{95.41}\\ p:96.10   r:94.73\end{tabular} \\[0.3cm]
$\surd$                      & $\surd$                      & $\surd$                       & $\surd$                          & $\surd$                         & $\surd$                      & $\surd$                      & $\surd$                         & $\surd$                           & $\surd$                           & & \begin{tabular}[c]{@{}c@{}}95.67\\ p:96.34   r:95.00\end{tabular} & \begin{tabular}[c]{@{}c@{}}97.67\\ p:97.46   r:97.88\end{tabular} & \begin{tabular}[c]{@{}c@{}}\textbf{96.03}\\ p:96.54   r:95.53\end{tabular}\\[0.3cm]

 $\surd$                      & $\surd$                      & $\surd$                       & $\surd$                          & $\surd$                         & $\surd$                      & $\surd$                      & $\surd$                         & $\surd$                           & $\surd$                           & $\surd$ & \begin{tabular}[c]{@{}c@{}}96.01\\ p:95.79   r:96.23\end{tabular} & \begin{tabular}[c]{@{}c@{}}98.14\\ p:97.32   r:98.98\end{tabular} & \begin{tabular}[c]{@{}c@{}}\textbf{96.33}\\ p:96.81   r:95.85\end{tabular} \\[0.3cm]
\hline
\end{tabular}
\caption{Experimental results of baseline, DTE (double training epoch), LCS (larger crop size), Flip (training and testing time flip), SyBN (SyncBN), DCN (deformable convolution network), LBS (larger batch size), Ranger optimizer, Reg 24 (a hyperparameter regmax), FPN (2-6) (FPN selection), WBF (weighted box fusion).}
\label{tab:ablation_study}
\end{table}

According to Table~\ref{tab:ablation_study}, we have the following observations.
\begin{itemize}[leftmargin=.3in]
  \item The number of training epochs should be large to make the model learn the data set fully.
  \item Large input resolution is good for the MFD task. Random crops from the images will cause the long-isolated formula to be truncated, and thus decrease the performance.
  \item ResNeSt is better than ResNet in this task. SyncBN can improve the performance a little. 
  \item DCN is a powerful weapon in detection tasks. DCN largely improves the performance of the isolated formulas. Such improvement can be attributed to that DCN can extend the receipt field of the network. That is important for large isolated formulas.
  \item Ranger optimizer has outperformed Adam optimizer consistently. Similar observation is also found in our another report~\cite{he2021ICDAR} about ICDAR 2021 Competition on Scientific Table Image Recognition to LaTeX~\cite{ye2021pingan}. In our evaluation of standard benchmarks, we also find that Ranger can improve the average accuracy by around 0.6\%. 
  \item The appropriate FPN level should be selected according to the scale of the object in the task. It can largely improve the performance.
  \item Weighted Box Fusion is useful in our task. It will improve the recall largely although it will slightly decrease the precision. Overall, it can improve the performance by 0.3\%.
\end{itemize}

{\bf{Remarks.}} Besides of the up-mentioned methods, we also tried many other approaches including  CutMix data augmentation, soft NMS instead of the original NMS, IoU loss and GIoU loss instead of the smooth L1 loss, BFP and PAFPN instead of the original FPN, and Swish and Mish activation functions instead of ReLU. However, unfortunately, we did not observe obvious improvement on our validation set.

\subsection{Final Results}
The final results are shown on Table~\ref{final_result} reported by the organizers, and our team is PAPCIC (Visual Computing Group from Ping An Property \& Casualty Insurance Company, China). The types, E, I, S characters, stand for Embedded, Isolated, and the Averaged, respectively. F1 score is used as the metric. Our method is ranked 1st place among all 15 teams. From Table~\ref{final_result}, we found that \begin{itemize}[leftmargin=.3in]
  \item almost all submissions obtained high performance on the isolated formula, but some methods performed poor on the embedded formula. The gap among the highest and the lowest on the isolated formula was around 3.0\%, but this gap amounted to  around 10.6\% on the embedded formula. It means most algorithms can detect the isolated formulas well, and the main challenge of the MFD task lies on the detection of the embedded formula.
  \item compared to the ``DLVCLab'' and ``TYAI'' methods, our method slightly beat them on the isolated formula, but outperformed them largely on the embedded formula. We believe this improvement is attributed to the large input resolution, the ATSS and the proper FPN choice. 
\end{itemize}

\begin{table}[]
\begin{tabular}{lcccc}
\multicolumn{1}{c}{Group ID} & Type                                              & \begin{tabular}[c]{@{}c@{}}F1\\ (Ts10+Ts11)\end{tabular}      & \begin{tabular}[c]{@{}c@{}}F1\\ Task dependent(Ts11)\end{tabular} & \begin{tabular}[c]{@{}c@{}}F1\\ Task independent(Ts10)\end{tabular} \\
\hline
PAPCIC                       & \begin{tabular}[c]{@{}c@{}}E\\ I\\ S\end{tabular} & \begin{tabular}[c]{@{}c@{}}94.89\\ 98.76\\ \bf{95.47}\end{tabular} & \begin{tabular}[c]{@{}c@{}}95.11\\ 98.70\\ \bf{95.68}\end{tabular}     & \begin{tabular}[c]{@{}c@{}}94.64\\ 98.79\\ \bf{95.37}\end{tabular}       \\
\hline
Lenovo                       & \begin{tabular}[c]{@{}c@{}}E\\ I\\ S\end{tabular} & \begin{tabular}[c]{@{}c@{}}94.29\\ 98.19\\ \bf{94.96}\end{tabular} & \begin{tabular}[c]{@{}c@{}}93.98\\ 97.85\\ \bf{94.60}\end{tabular}     & \begin{tabular}[c]{@{}c@{}}94.44\\ 98.33\\ \bf{95.13}\end{tabular}       \\
\hline
DLVCLab                      & \begin{tabular}[c]{@{}c@{}}E\\ I\\ S\end{tabular} & \begin{tabular}[c]{@{}c@{}}93.79\\ 98.54\\ \bf{94.60}\end{tabular} & \begin{tabular}[c]{@{}c@{}}93.88\\ 98.61\\ \bf{94.64}\end{tabular}     & \begin{tabular}[c]{@{}c@{}}93.75\\ 98.51\\ \bf{94.59}\end{tabular}       \\
\hline
TYAI                         & \begin{tabular}[c]{@{}c@{}}E\\ I\\ S\end{tabular} & \begin{tabular}[c]{@{}c@{}}93.39\\ 98.55\\ \bf{94.28}\end{tabular} & \begin{tabular}[c]{@{}c@{}}93.94\\ 98.42\\ \bf{94.66}\end{tabular}     & \begin{tabular}[c]{@{}c@{}}93.13\\ 98.61\\ \bf{94.10}\end{tabular}       \\
\hline
SPDBLab                      & \begin{tabular}[c]{@{}c@{}}E\\ I\\ S\end{tabular} & \begin{tabular}[c]{@{}c@{}}92.80\\ 98.06\\ \bf{93.70}\end{tabular} & \begin{tabular}[c]{@{}c@{}}92.14\\ 97.76\\ \bf{93.03}\end{tabular}     & \begin{tabular}[c]{@{}c@{}}93.12\\ 98.19\\ \bf{94.01}\end{tabular}       \\
\hline
YoudaoAI                     & \begin{tabular}[c]{@{}c@{}}E\\ I\\ S\end{tabular} & \begin{tabular}[c]{@{}c@{}}92.73\\ 98.34\\ \bf{93.70}\end{tabular} & \begin{tabular}[c]{@{}c@{}}92.71\\ 98.38\\ \bf{93.63}\end{tabular}     & \begin{tabular}[c]{@{}c@{}}92.74\\ 98.32\\ \bf{93.74}\end{tabular}       \\
\hline
PKUF-MFD                     & \begin{tabular}[c]{@{}c@{}}E\\ I\\ S\end{tabular} & \begin{tabular}[c]{@{}c@{}}91.94\\ 96.56\\ \bf{92.72}\end{tabular} & \begin{tabular}[c]{@{}c@{}}92.32\\ 96.87\\ \bf{93.04}\end{tabular}     & \begin{tabular}[c]{@{}c@{}}91.76\\ 96.43\\ \bf{92.57}\end{tabular}       \\
\hline
HW-L                         & \begin{tabular}[c]{@{}c@{}}E\\ I\\ S\end{tabular} & \begin{tabular}[c]{@{}c@{}}90.53\\ 98.94\\ \bf{91.97}\end{tabular} & \begin{tabular}[c]{@{}c@{}}90.57\\ 98.61\\ \bf{91.86}\end{tabular}     & \begin{tabular}[c]{@{}c@{}}90.51\\ 99.08\\ \bf{92.02}\end{tabular}       \\
\hline
Komachi                      & \begin{tabular}[c]{@{}c@{}}E\\ I\\ S\end{tabular} & \begin{tabular}[c]{@{}c@{}}90.39\\ 98.57\\ \bf{91.79}\end{tabular} & \begin{tabular}[c]{@{}c@{}}89.69\\ 98.60\\ \bf{91.11}\end{tabular}     & \begin{tabular}[c]{@{}c@{}}90.72\\ 98.55\\ \bf{92.10}\end{tabular}       \\
\hline
AIG                          & \begin{tabular}[c]{@{}c@{}}E\\ I\\ S\end{tabular} & \begin{tabular}[c]{@{}c@{}}89.71\\ 95.95\\ \bf{90.75}\end{tabular} & \begin{tabular}[c]{@{}c@{}}89.19\\ 96.07\\ \bf{90.26}\end{tabular}     & \begin{tabular}[c]{@{}c@{}}89.95\\ 95.90\\ \bf{90.97}\end{tabular}       \\
\hline
PKUSG                        & \begin{tabular}[c]{@{}c@{}}E\\ I\\ S\end{tabular} & \begin{tabular}[c]{@{}c@{}}89.10\\ 97.96\\ \bf{90.62}\end{tabular} & \begin{tabular}[c]{@{}c@{}}88.59\\ 97.94\\ \bf{90.09}\end{tabular}     & \begin{tabular}[c]{@{}c@{}}89.34\\ 97.96\\ \bf{90.87}\end{tabular}       \\
\hline
TAL                          & \begin{tabular}[c]{@{}c@{}}E\\ I\\ S\end{tabular} & \begin{tabular}[c]{@{}c@{}}87.87\\ 96.85\\ \bf{89.42}\end{tabular} & \begin{tabular}[c]{@{}c@{}}88.51\\ 97.09\\ \bf{89.89}\end{tabular}     & \begin{tabular}[c]{@{}c@{}}87.57\\ 96.75\\ \bf{89.19}\end{tabular}       \\
\hline
UIT                          & \begin{tabular}[c]{@{}c@{}}E\\ I\\ S\end{tabular} & \begin{tabular}[c]{@{}c@{}}86.04\\ 97.05\\ \bf{87.94}\end{tabular} & \begin{tabular}[c]{@{}c@{}}85.64\\ 98.11\\ \bf{87.63}\end{tabular}     & \begin{tabular}[c]{@{}c@{}}86.23\\ 96.60\\ \bf{88.08}\end{tabular}       \\
\hline
AV-DFKI                      & \begin{tabular}[c]{@{}c@{}}E\\ I\\ S\end{tabular} & \begin{tabular}[c]{@{}c@{}}85.35\\ 97.48\\ \bf{87.45}\end{tabular} & \begin{tabular}[c]{@{}c@{}}84.75\\ 97.40\\ \bf{86.80}\end{tabular}     & \begin{tabular}[c]{@{}c@{}}85.63\\ 97.52\\ \bf{87.76}\end{tabular}       \\
\hline
VH                           & \begin{tabular}[c]{@{}c@{}}E\\ I\\ S\end{tabular} & \begin{tabular}[c]{@{}c@{}}84.25\\ 98.59\\ \bf{86.67}\end{tabular} & \begin{tabular}[c]{@{}c@{}}84.39\\ 98.51\\ \bf{86.61}\end{tabular}     & \begin{tabular}[c]{@{}c@{}}84.18\\ 98.62\\ \bf{86.70}\end{tabular}  \\
\hline 
\end{tabular}
\caption{Final results of all 15 teams on ICDAR 2021 Competition on Mathematical Formula Detection.}
\label{final_result}
\end{table}

\section{Conclusion}
\label{sec:conclusion}
In this paper, we presented our 1st place method to ICDAR 2021 Competition on Mathematical Formula Detection. We built our approach on GFL, with ResNeSt as our backbone, to detect embedded and isolated formulas. We employed ATSS as our sampling strategy instead of random sampling to eliminate the effects of sample imbalance. Moreover, we observed and revealed the influence of different FPN levels on the detection result. Finally, with a series of useful tricks and model ensembles, our method was ranked 1st in the MFD task.

\bibliographystyle{unsrt}  
\bibliography{references}

\end{document}